\documentclass[conference]{IEEEtran}
\IEEEoverridecommandlockouts
\usepackage{cite}
\usepackage{amsmath,amssymb,amsfonts}
\usepackage{graphicx}
\usepackage{textcomp}
\usepackage{xcolor}
\usepackage{placeins}
\usepackage{float}
\usepackage{url}

\begin{document}

\title{Agent Trust Infrastructure (ATI): A Federated Identity, Trust,
Attestation, and Accountability Architecture for Cross-Organization Autonomous
AI Agents}

\author{
\IEEEauthorblockN{Viet K. Nguyen}
\IEEEauthorblockA{\textit{California State Polytechnic University, Pomona} \\
Pomona, CA 91768, USA \\
vietknguyen@cpp.edu}
\and
\IEEEauthorblockN{Mohammad I. Husain}
\IEEEauthorblockA{\textit{California State Polytechnic University, Pomona} \\
Pomona, CA 91768, USA \\
mihusain@cpp.edu}
}

\maketitle

\begin{abstract}

AI agents increasingly collaborate across organizational boundaries,
invoking tools and interacting with agents operated by independent
entities. Although existing agent frameworks support planning and
coordination, they lack a common mechanism for verifying
cross-organizational agent identity, execution integrity, and
authorization. We propose the Agent Trust Infrastructure (ATI),
a framework-independent security architecture built on Public Key
Infrastructure (PKI), Zero Trust Architecture (ZTA), and remote
attestation. ATI establishes federated trust through peer
organizational certificate authorities (AgentCAs), eliminating
the need for a shared root authority. Before executing requests,
ATI validates agent certificates, verifies approved model, tool,
and system-prompt configurations, and enforces authorization
through three policy checks. 

We implement ATI as an external security enforcement layer that
integrates with existing agent frameworks without modifying their
underlying models. We evaluate ATI in a simulated federation of
eight organizations using six agent frameworks, each deploying
a five-agent team. Security experiments involve two organizations
and assess 14 attack scenarios and six benign tasks across five
agent models, using five attack variations and three repetitions
per variation. A panel of judge models evaluates the experimental
outcomes. With all enforcement mechanisms enabled, attack
containment increases from 43\% to 84\% (40\% to 80\% for the
11 attack scenarios evaluated consistently across all
configurations), while enforcement introduces
approximately 174~$\mu$s of overhead per request. ATI is released
as an open-source, framework-independent package that uses IETF
WIMSE workload token formats for identity and request verification
and Entity Attestation Tokens for model, tool, and system-prompt
integrity verification.
\end{abstract}


\section{Introduction}

AI agents differ fundamentally from traditional software services.
While conventional services execute predefined logic, autonomous
agents dynamically plan, maintain memory, invoke tools, collaborate
with other agents, and delegate tasks. As organizations increasingly
deploy agentic AI systems, interactions extend beyond institutional
boundaries to agents operated by external organizations and cloud
providers. In such environments, network-level trust and API-based
authentication alone are insufficient to establish whether an agent
is legitimate, operating as intended, or authorized to perform a
requested action.

We argue that autonomous AI agents must be treated as independent
security principals, similar to users, devices, and services.
Each agent requires a cryptographically verifiable identity,
cross-organizational trust, mechanisms to verify its approved
model and execution configuration, fine-grained authorization,
and auditable records of its actions. These capabilities must
operate independently of the agent's underlying model and
orchestration framework to support heterogeneous, cross-vendor
environments.

To address these requirements, we introduce the Agent Trust
Infrastructure (ATI), a framework-independent architecture
that combines certificate-based identity, federated trust,
runtime attestation, policy enforcement, and cryptographic
accountability. ATI places security enforcement outside the
agent's language model, allowing trust decisions to be made
through deterministic verification rather than relying on
the model's behavior.

The primary contributions of this work are:

\begin{itemize}

\item \textbf{A five-plane trust architecture:}
We design ATI around five complementary planes---Identity,
Trust, Attestation, Policy, and Accountability. Security
checks execute independently of the agent's model and
orchestration framework, enabling consistent enforcement
across heterogeneous agentic AI systems.

\item \textbf{Federated certificate-based agent identity:}
We introduce AgentCA, an organizational certificate authority
that establishes cryptographically verifiable agent identities.
Independent AgentCAs form peer trust relationships without
requiring a common root authority. Certificate issuance is
restricted to each authority's organizational namespace,
supporting decentralized administration and cross-organizational
identity verification.

\item \textbf{Runtime attestation of agent configurations:}
We introduce attestation mechanisms that verify an agent's
model, approved tools, and system prompt against trusted
configurations. Unlike certificate validation alone, these
checks detect unauthorized changes to agent configurations.
In our experiments, attestation prevented all tested model
substitution and system-prompt modification attacks.

\item \textbf{Incremental security evaluation:}
We evaluate ATI using 14 attack scenarios across five agent
models. An incremental evaluation enables successive trust
planes to be introduced and their contributions to attack
containment to be measured independently.

\item \textbf{Cross-model security analysis:}
We demonstrate that security outcomes vary substantially
across agent models, with identical attacks rejected by some
models but executed by others. In contrast, ATI's deterministic
entry-point checks consistently prevent their targeted attacks
across nearly all trials, demonstrating the value of
model-independent security enforcement.

\item \textbf{Open-source, framework-independent implementation:}
We release \texttt{ati}, an open-source package ~\cite{husain2026ati} providing
a command-line interface and security enforcement proxy
that can be deployed in front of existing agents without
modifying their underlying models. Unlike the agent governance
solutions examined in this study, ATI enables receiving
organizations to verify the calling agent's model, tool
configuration, and system prompt.

\end{itemize}

\section{Why Existing Trust Primitives Do Not Fit Agents}
Current agent frameworks coordinate agents but do not establish trust between
them. CrewAI,
AutoGen, LangGraph, the OpenAI Agents SDK, Anthropic MCP, Semantic Kernel, and
Google ADK all provide planning, memory, tool invocation, and multi-agent
coordination, and each defines agents in its own types. None of them gives an
agent an identity that another framework, let alone another organization, can
verify. Inside one deployment this does not matter, because every agent is
trusted by default. It matters as soon as a request arrives from another
organization.

The identity technologies that would normally fill that gap were built for other
principals. PKI and X.509 bind names to keys for hosts and people. OAuth 2.0 and
OpenID Connect authorize applications that act for a user. SPIFFE/SPIRE issues
short-lived identities to workloads. Decentralized identifiers and verifiable
credentials provide portable, issuer-signed claims. TPM-based attestation and
trusted execution environments prove what code is running. The NIST Zero Trust
architecture adds the rule that every request is checked where it arrives. Each
of these works well for what it was built for. None was built for an agent,
which hands off work on its own and whose behavior depends on which model and
system prompt it is running at the moment.

We found no design that puts the existing building blocks together: a checkable
agent identity, trust across organizations, attestation of the model and prompt,
permission checks on each tool call, and a tamper-evident record, independent of
the agent framework.

\section{Related Work}
The trust work for AI agents has grown rapidly, but most of it addresses only partial aspects
of the problem.

\textbf{Agent identity and credentials.} Several proposals give an agent a
first-class, verifiable identity. OpenID Connect for Agents (OIDC-A) extends OIDC
with agent claims, delegation chains, and attestation claims~\cite{oidca}.
Other work equips agents with decentralized identifiers and verifiable
credentials~\cite{didvc}, or proposes a zero-trust identity framework with
decentralized authentication and fine-grained access
control~\cite{zerotrustid}. An OpenID Foundation whitepaper maps the emerging
identity-management landscape for agentic AI~\cite{idmgmt}. These efforts define what an agent identity looks
like and how it is issued. In ATI, identity is one of five planes, and each
organization's certificate authority (AgentCA) issues certificates that tie an
agent's identifier to its organization and are checked on every request.

\textbf{Delegation and authorization.} South et al. propose authenticated,
authorized, and auditable delegation, extending OAuth 2.0 and OIDC so a user can
give an agent limited, revocable authority with a clear chain of
accountability~\cite{authdeleg}. Registry-based approaches defend multi-agent
systems against tool squatting with a zero-trust tool registry~\cite{toolsquat}.
ATI's Policy plane instead checks permissions at every tool call (writes to
other organizations, repeated transactions, and bulk reads), not only when a
credential is issued.

\textbf{Zero trust for multi-agent systems.} A recent survey reviews
zero-trust design for multi-LLM systems at the edge, covering strong
identification, context-aware access control, and blockchain-based
management~\cite{ztsurvey}. The never-trust-by-default stance is written down for
enterprise networks in NIST SP 800-207~\cite{nist80207}, and SPIFFE issues
cryptographically attested identities to workloads rather than
users~\cite{spiffe}. ATI applies the same idea to agents. It puts all the checks
in one reference monitor, a single checkpoint that every request must pass, which
runs outside the agent's model, and we measure how much each check helps.

\textbf{Interoperability protocols and their security.} Anthropic's Model Context
Protocol (MCP)~\cite{mcp} and Google's Agent2Agent (A2A)~\cite{a2a} standardize
how agents reach tools and each other; surveys compare these with ACP and
ANP~\cite{interopsurvey}. Security work on these protocols includes threat
models for MCP, A2A, and related protocols~\cite{protothreat}, analysis of tool
squatting, deceptive registration of tools~\cite{toolsquat}, and a documented remote code execution
flaw in an MCP component~\cite{mcpcve}. ATI does not depend on the protocol. Its
checks run in code where a request enters, so the same planes apply to whichever
protocol carries the request.

\textbf{Runtime integrity, prompt injection, and attestation.} Prompt injection
is a persistent threat for agents that read untrusted content, with dedicated
benchmarks such as AgentDojo~\cite{agentdojo} and a broad survey of agentic
threats and defenses~\cite{agenticsec}. Confidential computing uses hardware
attestation to check the environment in which a model runs~\cite{ccsurvey}. ATI's
Attestation plane checks the running model, the tool list, and a hash of the
agent's system prompt against the approved versions. It does not defend against
instructions hidden in the data that a request carries, which we treat as a separate
problem.

\textbf{Accountability and non-repudiation.} BAID binds an agent to its
operator's identity and proves which program it runs with zero-knowledge proofs
over the program binary, for accountability and code integrity~\cite{bindingid}. ATI's Accountability plane records every allowed
action in a signed, hash-chained log that ties an action to the acting
identity and model.

\textbf{Open-source agent governance tools.} Several installable tools cover
parts of this problem, including Microsoft's Agent Governance Toolkit
(AGT)~\cite{agt}, OrgKernel~\cite{orgkernel}, and AGNTCY
Identity~\cite{agntcyid}, and the Agent Name Service proposes a registry to
find and verify agents~\cite{ans}. 

\textbf{Positioning.} Prior work tends to cover a single plane (identity,
delegation, or accountability), or proposes a standard without testing it across
frameworks. ATI combines all five planes into one reference monitor outside
 the agent framework. ATI also does not add no new cryptography. Each plane reuses a known design: X.509
certificates with name constraints, short-lived workload identity as in SPIFFE,
the measure-then-attest pattern of TPMs with the model, tools, and prompt as 
measured, a zero-trust reference monitor~\cite{nist80207}, and a signed 
append-only log like those used for certificate transparency.

This work adds two distinct features that we did not find in the open-source tools above,
checked against their public repositories on October 15, 2026.

\textbf{The receiving organization checks what the caller runs.} The caller's
own CA writes the approved model, tool list, and system-prompt hash into the
agent's identity token. When the token carries this record, every request must carry a signed report
of what the agent is running now, which the receiver compares against that record. In the tools we
examined, checks on an agent's model and tools, where they exist, stay inside
the organization that runs the agent, and we found no way for another
organization to see or check them. This is what stops three attacks in Section~\ref{sec:eval}
that pass every identity check: a swapped model (A7), a tampered tool (A8), and an
edited prompt (A14), since the agent's key and certificate do not change.

\textbf{ATI is measured against attacks.} We report how often each attack
succeeds with and without each plane, across five agent models. AGT reports detection
rates for its prompt-injection detector, but states that it has not yet published
an attack-success benchmark for its enforcement layer~\cite{agt}, and we found
none for the other tools. Such numbers matter because the same attack is refused by one model and carried out
by another, so a tool tested on one model can look safer than it is.

\section{System Requirements and Threat Model}\label{sec:threat}
\begin{figure*}[t]
  \centering
  \includegraphics[width=\textwidth]{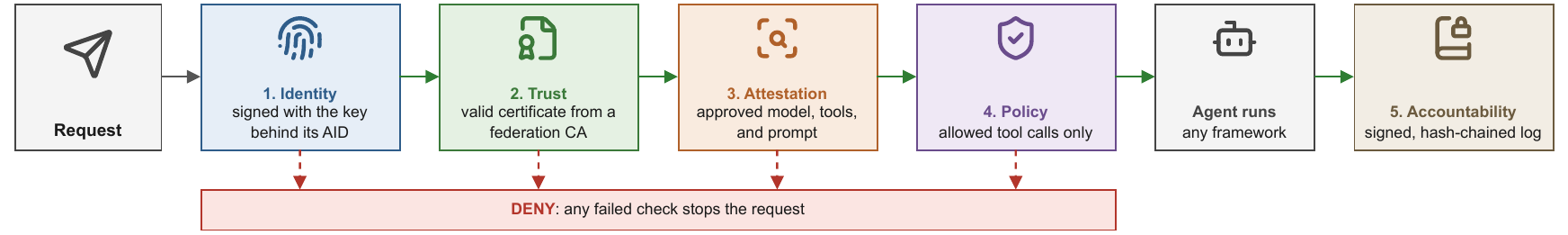}
  \caption{The five ATI planes. Identity, Trust, and Attestation check a request
  before the agent runs, and Policy checks each tool call the agent makes. Any of
  them can deny it. Accountability records each allowed action.}
  \label{fig:overview}
\end{figure*}
We treat each AI agent as an independent principal that can start
conversations, call tools, hand off work, and exchange information with other
organizations.

Four requirements follow. An agent needs an identifier that another organization can
verify without prior contact. Trust must cross organizational boundaries without
a common authority above them. The model and prompt actually running must be
checkable, since an agent's behavior depends on both. And every action must be
attributable afterward to the identity and the model that produced it.

The threat model considers external attackers, compromised agents, malicious
insiders, stolen certificates, replay attacks, prompt injection, model
replacement, dependency tampering, unauthorized delegation, supply-chain attacks,
collusion among compromised agents, expired and cross-organization certificates,
and tampering of an agent's own system prompt. We want every request to come
from a known agent, every action to be allowed, every agent to run its approved
setup, and every action to be traceable afterward.
Encrypting traffic between organizations and keeping the system available under
load are outside the scope of this work.

\section{Agent Trust Infrastructure (ATI)}
ATI is organized as five planes. Identity, Trust, and Attestation check a request
before the agent's model runs, and Policy checks each tool call the agent then
makes. Any of them can deny it (Fig.~\ref{fig:overview}).

\subsection{Identity Plane}
\begin{figure}[t]
  \centering
  \includegraphics[width=\columnwidth]{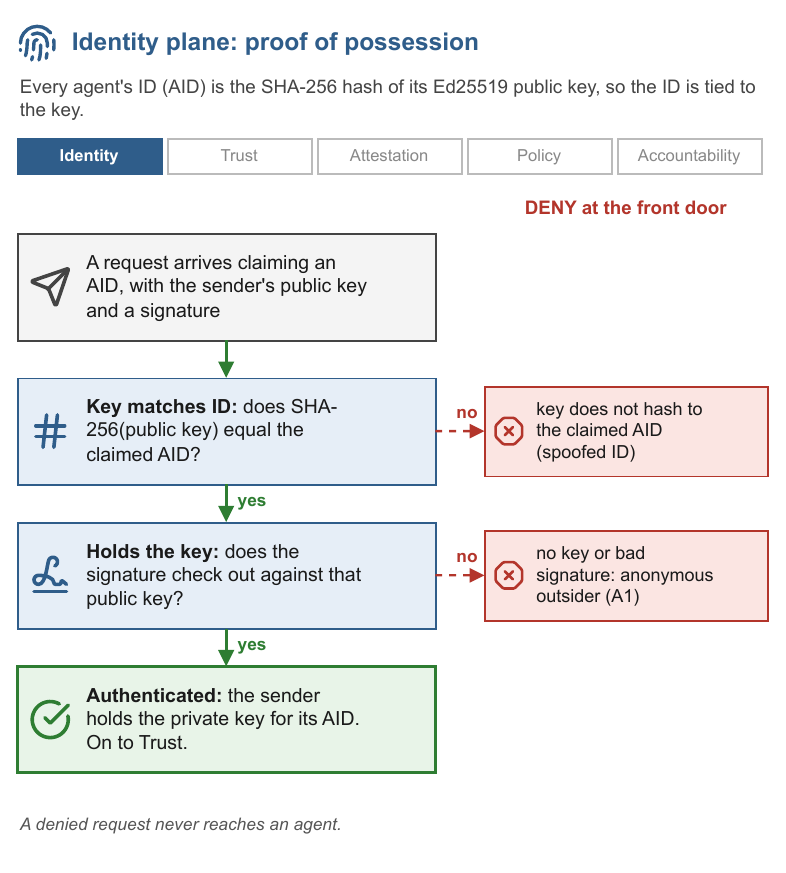}
  \caption{The Identity plane. A request with no key is refused at the front door. A keyed request is signed with its key, and its AID is the hash of that key; the Trust plane then checks that the certificate names this AID.}
  \label{fig:identity}
\end{figure}
The Identity Plane gives every agent a globally unique Agent Identifier (AID) and
checks that each request is signed with the agent's private key. Each agent
holds an Ed25519 key pair, and its AID is the SHA-256 hash of its public key. The
identifier is therefore tied to the key and cannot be picked freely. (The
simulation shortens the hash to 64 bits; a deployment would use all of it.) To
authenticate, an agent signs the request with its private key, and the plane
checks the signature against the public key. A request with no key is anonymous and is
refused at the organization's front door before any agent runs (Fig.~\ref{fig:identity}).
This mirrors
cryptographic workload identity such as SPIFFE~\cite{spiffe} and recent
agent-identity proposals that give agents verifiable identifiers~\cite{oidca,didvc}.

This stops the outside attacker with no credentials (A1). With no defense, this
attack is stopped only 20\% of the time, and only when the model itself happens
to refuse. Once identity is required, it is stopped every time.

We keep the plane small on purpose. It works together with the Trust Plane to
stop impersonation. An attacker who signs with their own key does not match the
AID in the certificate (Section~\ref{sec:trust}), and an attacker who has a
certificate but not its private key cannot produce a valid signature. Because the
simulation runs in one process, the signature is created and checked in the same
place and never crosses a network; a real deployment would send it with the
request. Our claim about a certificate used without its key therefore rests on
this argument and was not measured. We also leave out key rotation and other key
management, since no attack in our threat model needs it.

\subsection{Trust Plane: AgentCA and Federation}\label{sec:trust}
\begin{figure}[t]
  \centering
  \includegraphics[width=\columnwidth]{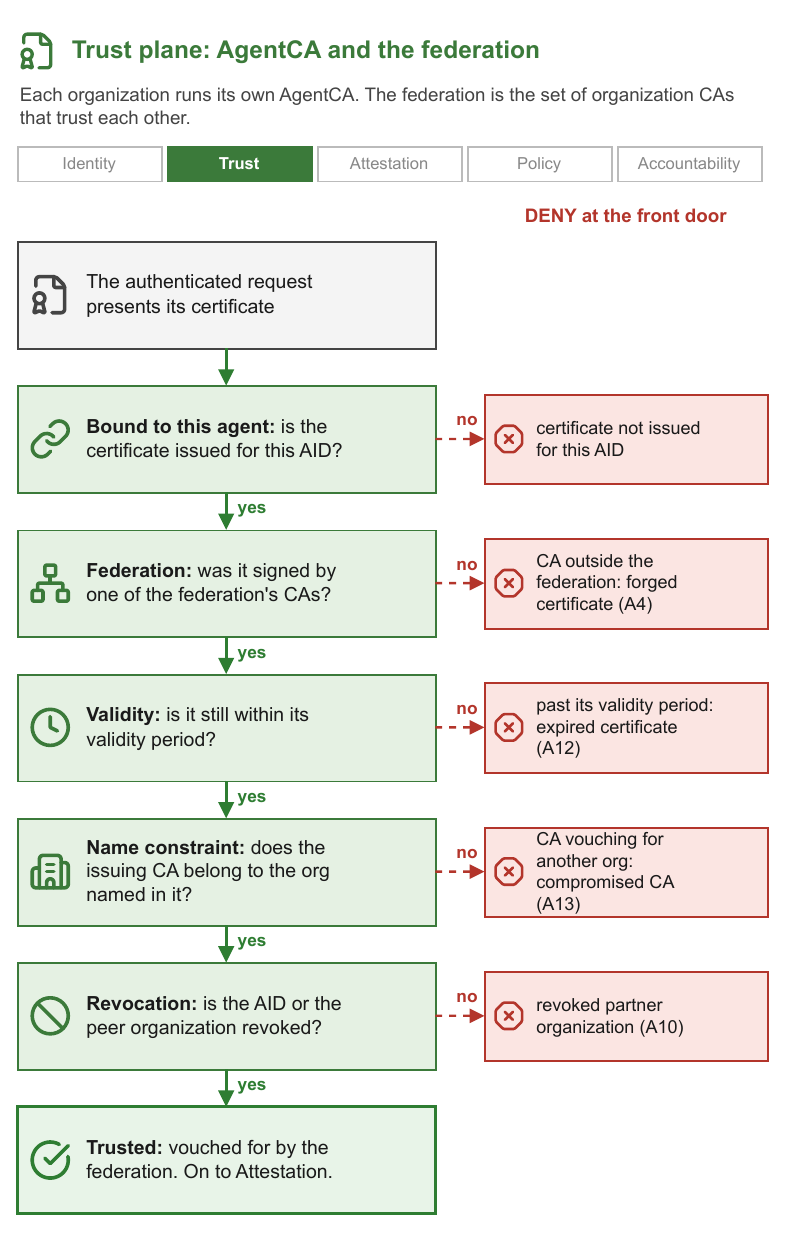}
  \caption{The Trust plane. The certificate must be issued for the sender's AID, signed by a federation CA that vouches only for its own organization, unexpired, and not revoked. Any failed check refuses the request.}
  \label{fig:trust-detail}
\end{figure}
Identity establishes who is calling. The Trust Plane checks whether that
identity is vouched for by a certificate authority the federation accepts. Each
organization runs an AgentCA, which issues certificates that tie an AID to an
organization and a role and signs them with the CA's key, following
X.509~\cite{x509}. The eight organizations' CAs form a federation, and a
certificate is trusted only if one of them signed it.

A request passes the Trust Plane only if all of the following hold. The
certificate is signed by a federated CA; the signature verifies; the
certificate's AID matches the key that signed the request; the CA that issued it
belongs to the organization named in the certificate; the certificate is within its validity
window; and the AID is not revoked (Fig.~\ref{fig:trust-detail}). Our
simulation tests only the revocation of a whole partner organization (A10).

\textbf{Certificate fields and hierarchy.} Other proposals put many descriptive
attributes in an agent's credentials. OIDC-A, for instance, defines claims for
the agent's provider, model, capabilities, and trust level~\cite{oidca}, and
delegation proposals add signed tokens with a scope and an
expiry~\cite{authdeleg}. To issue certificates at scale, PKI usually uses a
hierarchy of a root CA, organization CAs, and registration authorities. We carry
only the fields that some check actually reads: AID, organization, role, issuer,
validity period, and signature. No check in our threat model reads the other
attributes, so including them would not change any result. We also use a flat
federation instead of a hierarchy. The main security reason for a hierarchy is to
make sure each CA vouches only for its own organization. We enforce that directly
with a name constraint, which requires the issuer to match the certificate's
organization. The root and registration-authority layers handle vetting and
onboarding when certificates are issued, which happens before any request is
checked.

\textbf{Attacks closed.} The Trust Plane stops the stolen or forged certificate
(A4). A certificate signed by an authority outside the federation passes the
Identity check, because the key is real, but fails the Trust check. The share of
A4 attempts stopped rises from 56\% with Identity alone to 100\%. The plane also
stops the compromised partner organization (A10) by refusing to talk to a revoked
peer, raising the rate from 37\% to 93\%. It also stops two certificate attacks we
added to the threat set. An expired but otherwise valid certificate (A12) is
accepted by the baseline, which never checks dates, and is stopped only 54\% of
the time, when the model happens to refuse; once expiry is checked it is stopped
every time. A certificate in which a federation CA vouches for an organization it
does not own (A13), as a compromised CA might issue, is stopped 75\% of the time
with no defense and every time once the name constraint is checked.

\textbf{Fidelity.} The certificates are real Ed25519 signatures over their fields,
checked on every request. The expiry date is currently stored next to the signed
data rather than inside it. A full deployment would sign it too; this does not
change any result here.

\subsection{Attestation Plane}
\begin{figure}[t]
  \centering
  \includegraphics[width=\columnwidth]{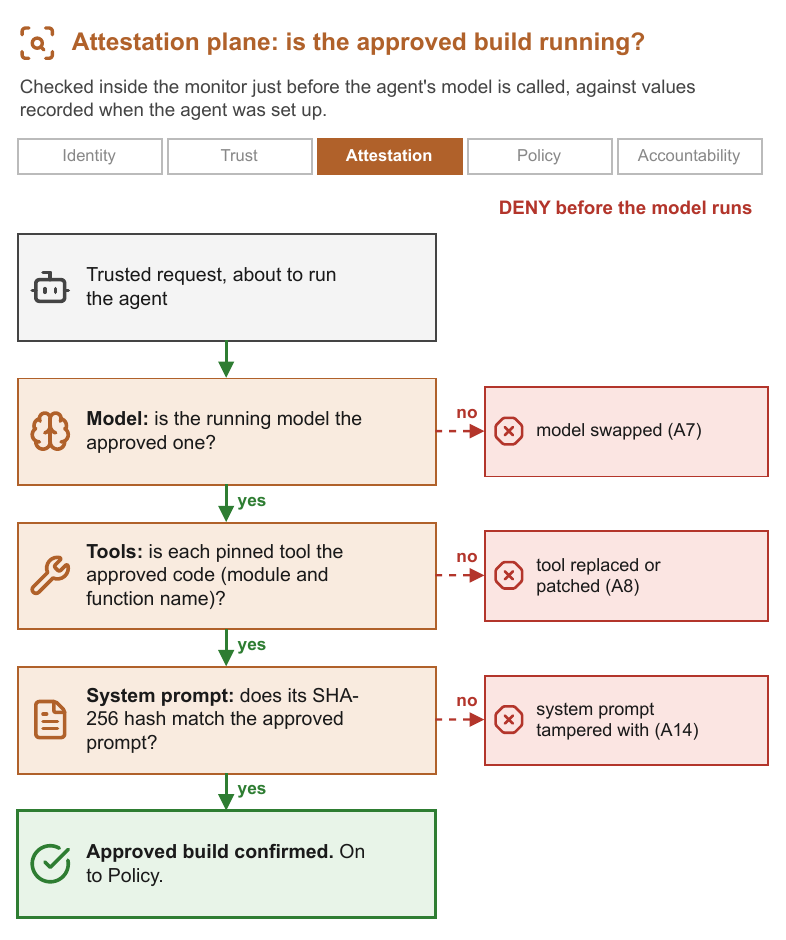}
  \caption{The Attestation plane. Before the agent runs, its model, its pinned tools, and a hash of its system prompt must match the approved versions. Any mismatch is refused.}
  \label{fig:attestation}
\end{figure}
Identity and Trust tell us who is calling. They say nothing about what is running
behind that identity. The Attestation Plane checks the agent's setup before it
acts and stops if it does not match the approved one.

The plane checks three things. It compares the running model to the approved
model, the pinned tool (\texttt{web\_search}) to its approved code (by module and
function name), and a hash of the agent's system prompt to the approved prompt. A substituted model, a tampered tool, or a swapped system
prompt no longer matches and is rejected before the model is invoked
(Fig.~\ref{fig:attestation}).

These checks stop model replacement (A7), where a different model is quietly
swapped in, raising the stop rate from 0\% to 100\%. They also stop dependency
tampering (A8), where a tool is replaced with a poisoned version, raising it from
81\% to 100\%. Checking the prompt stops system-prompt tampering (A14), a case we
added. It differs from prompt injection (A6), which hides instructions in data
the agent reads~\cite{agentdojo,agenticsec}; in A14 the agent's own instructions
are replaced. With no defense, the tampered prompt is followed and the attack
is stopped only 27\% of the time, when the model refuses. Checking the prompt as well stops it every time.

\textbf{Scope and fidelity.} A full attestation stack would also measure the
binary, the container image, and the host environment, and would let a remote
party check those measurements before trusting the agent, as remote attestation
does for confidential computing~\cite{ccsurvey}. We do not measure binaries,
containers, or the host, which have no representation in a code-level simulation.
Our checks are simple
comparisons and hashes done in the same process, not signed reports. A signature
would add nothing here, because no separate, untrusted party produces the
measurement. In our simulation the check runs once, when a request first enters
the federation, not on a timer. Agents the request reaches later, in another
office or another organization, are not checked again (Section~\ref{sec:crossorg}).

\subsection{Policy Plane}
\begin{figure}[t]
  \centering
  \includegraphics[width=\columnwidth]{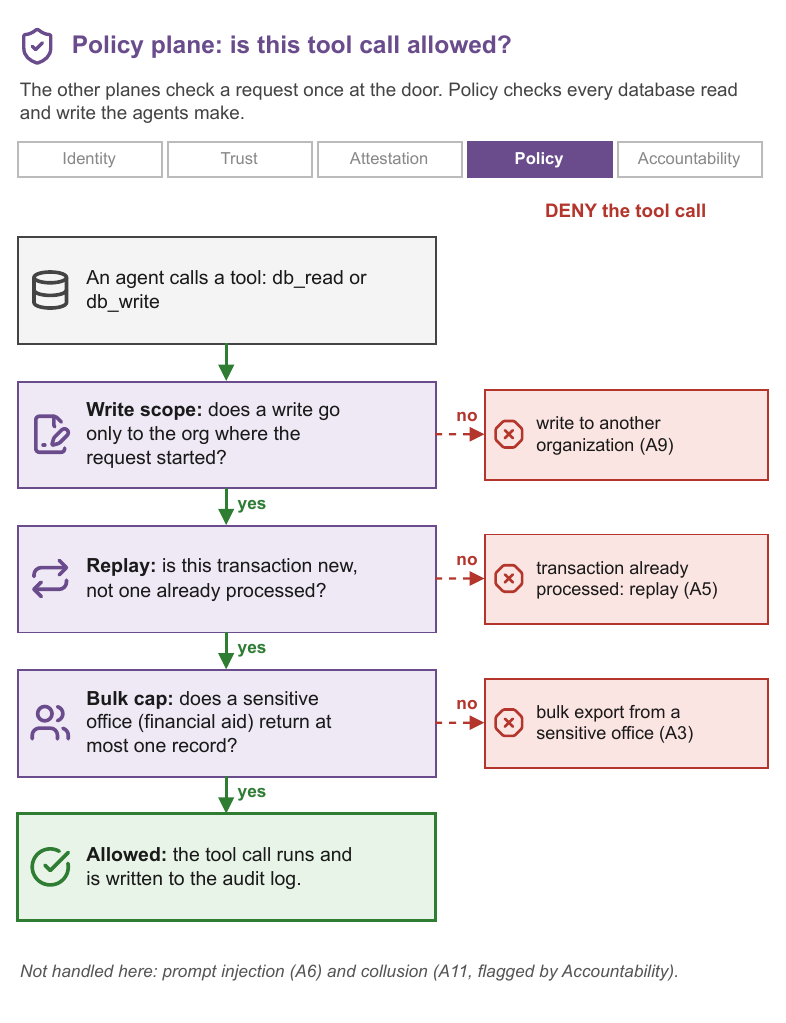}
  \caption{The Policy plane. Every database read and write is checked for writes to another organization, replayed transactions, and bulk exports from a sensitive office. Prompt injection (A6) and collusion (A11) are not handled here.}
  \label{fig:policy}
\end{figure}
The first three planes tell us who is calling and that the agent runs its
approved setup. The Policy Plane decides whether that caller may take a specific
action. The earlier planes check a request once, at the organization's front
door. The Policy Plane checks every tool call, meaning each database read and
write, so permissions are checked throughout the work the agents do. This complements proposals that limit what a delegated credential
grants~\cite{authdeleg} and registries that defend against deceptive
tools~\cite{toolsquat}.

The plane enforces three rules. First, organizations can read each other's
data but not write to it, so a write to any organization other than the one where
the request started is refused. This mostly stops unauthorized cross-organization
delegation (A9), raising the stop rate from 16\% to 71\%. Second, a transaction
already marked as processed is not processed again, which raises the stop rate
for replay attacks (A5) from 15\% to 75\%. Third, a sensitive office (financial
aid) refuses any read that would return more than one record, a backstop
against the malicious insider's bulk export (A3) (Fig.~\ref{fig:policy}); A3 is
already refused by most models, so the table shows no measurable gain from this
rule. These
rules run in code at the tool call, so when the agent makes the forbidden call,
the answer does not depend on the model or the wording. That is why A5 and A9
jump once the plane is on. They stop short of 100\% because a rule fires only on
the call it matches: the cross-organization rule fires only if the agent calls
\texttt{db\_write}, and the replay rule looks for the transaction ID in the
write, which some models leave out.

\textbf{Delegation.} Delegation is usually done with signed tokens that carry a
scope, an expiry, and permissions, so the agent receiving a task can check what
the sender was allowed to hand off~\cite{authdeleg,oidca}. Our simulation does not
need these tokens, because one reference monitor sees every call across offices
and organizations and can check a delegated action against where it started.
Tokens do the same check in a distributed system, where the receiving service
cannot see the sender's context; with a single monitor they would not change any
decision. We therefore enforce the one delegation rule our threat model tests,
that a request from another organization cannot cause a write, through the
origin rule above, and leave tokens to a real deployment. Finer-grained,
role-based permissions that would catch privilege escalation inside one
organization are not part of ATI yet.

\subsection{Accountability Plane}\label{sec:acct}
\begin{figure}[t]
  \centering
  \includegraphics[width=\columnwidth]{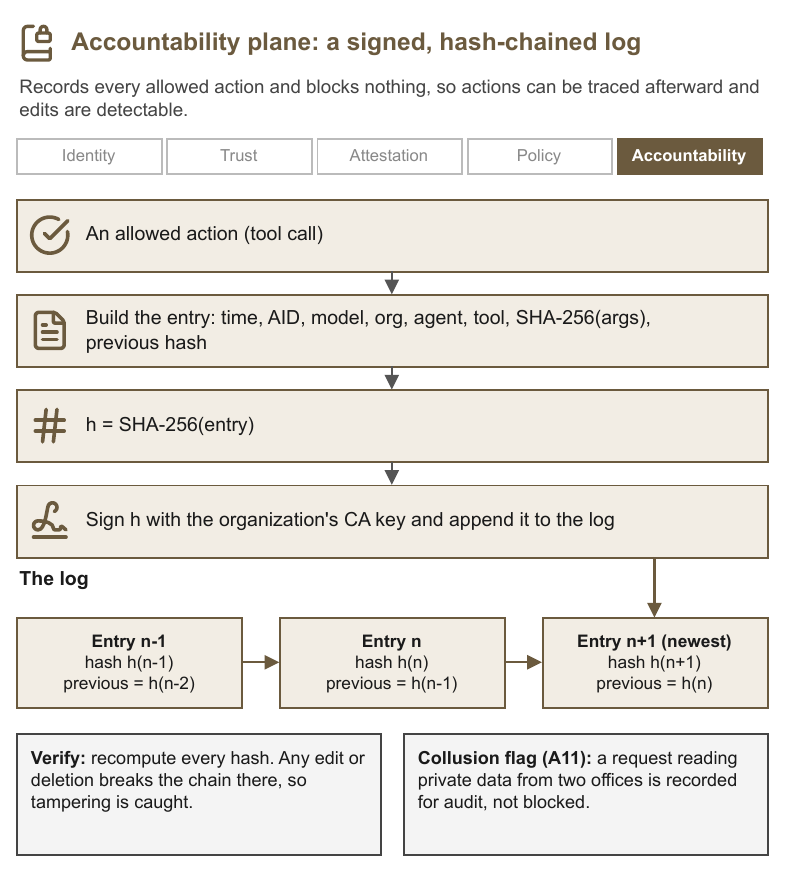}
  \caption{The Accountability plane. Each allowed database read and write becomes a signed log entry that includes the hash of the entry before it, so any later edit breaks the chain. A request that reads private data from two offices (A11) is flagged for review.}
  \label{fig:accountability}
\end{figure}
The four earlier planes block requests. The Accountability Plane blocks nothing.
It records every checked action so that each one can be traced afterward to who
did it, the accountability goal that recent work sets for
agents~\cite{bindingid}. It keeps an append-only log in which each entry records
the agent that made the request (its AID), the model that acted, a timestamp, the
tool called, and a hash of the tool's arguments. Each entry includes the hash of
the entry before it and is signed with the organization's CA key, and the
signature covers the identity, model, and timestamp. Any later edit, or deletion of an entry in the middle,
breaks the chain, and a separate \texttt{verify} step detects it
(Fig.~\ref{fig:accountability}).

This gives two properties. First, an agent cannot later deny an action that is in the log
(non-repudiation), as long as the organization's signing key was not stolen. Each entry names the agent that caused the action and the model
that carried it out, not just the office involved, and cannot be quietly changed.
Second, the log shows what happened without exposing content, because arguments
are stored as hashes rather than raw text. Someone can check the record without
seeing sensitive prompts or records.

The plane also catches one pattern that the blocking planes allow on purpose.
In the collusion case (A11), one request reads private data from two offices to
build a combined profile. A legitimate multi-office task looks exactly the same,
so a rule that blocked it would also block normal work. Instead, the log flags any
single request that reads private data from two or more offices as a
separation-of-duties violation, so it can be reviewed later rather than refused.

We record only the fields an auditor can act on and the simulation can fill in
accurately: identity, model, timestamp, tool, and argument hash, all signed.
Hashes of the prompt, the model's reasoning, and its output, and a record of who
delegated to whom, would fit in the same entry but are not captured yet.

\subsection{Working Across Organizations}\label{sec:crossorg}
\begin{figure}[t]
  \centering
  \includegraphics[width=\columnwidth]{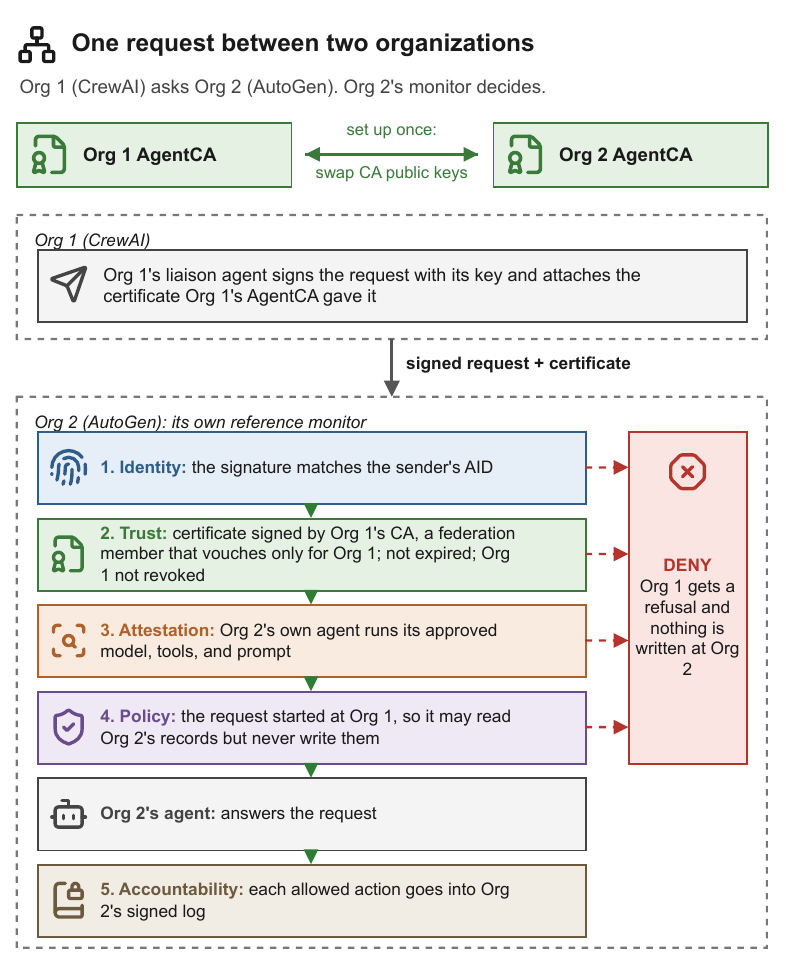}
  \caption{One request between two organizations. The organizations share only
  their CA public keys. The receiving organization's monitor checks the sender's
  identity and certificate, its own agent's build, and what the request may do,
  and records each allowed action in its own log.}
  \label{fig:crossorg}
\end{figure}
ATI is meant for agents that work across organizations, so the planes have to
fit together between two of them. Fig.~\ref{fig:crossorg} follows one request
from Org~1, whose agents run on CrewAI, to Org~2, whose agents run on AutoGen.
Before any request, the two organizations swap their AgentCA public keys. That is
all they share. Each one still issues certificates only to its own agents and
keeps its own approved builds, policy rules, and log.

When Org~1's liaison agent needs something from Org~2, it signs the request with
its private key and attaches the certificate that Org~1's AgentCA issued it.
Org~2's reference monitor checks the request like any other. Identity checks that
the signature matches the sender's AID. Trust checks that the certificate was
signed by Org~1's CA, that this CA is in the federation and is vouching for its
own organization, that the certificate has not expired, and that Org~2 has not
revoked Org~1. Org~2 then checks its own side. Attestation confirms that the
Org~2 agent about to answer runs its approved model, tools, and prompt.
Policy lets a request that started at Org~1 read Org~2's records but never write
them. Each allowed action goes into Org~2's log. If any check fails, Org~1 gets a
refusal and nothing is written at Org~2. Neither organization has to trust the
other's framework or model, because each decision is made in code by the
organization that owns the data. If Org~1 is later compromised, Org~2 revokes it
and its requests are refused (A10).

Our simulation simplifies this, because all eight organizations run in one
process. The requester's key and certificate are checked once, where the request
enters the federation, and so is the attestation of the agent it reaches first.
At the hop from one organization's liaison to another, the only entry check is
whether the partner has been revoked; Policy still checks every tool call at the
receiving organization. All organizations write to one shared log, with each
entry signed by the key of the organization where the action happened. A
deployment would send the signature and certificate with every hop, attest the
receiving agent there, and keep one log per organization.

\section{Reference Implementation}
\begin{figure*}[t]
  \centering
  \includegraphics[width=\textwidth]{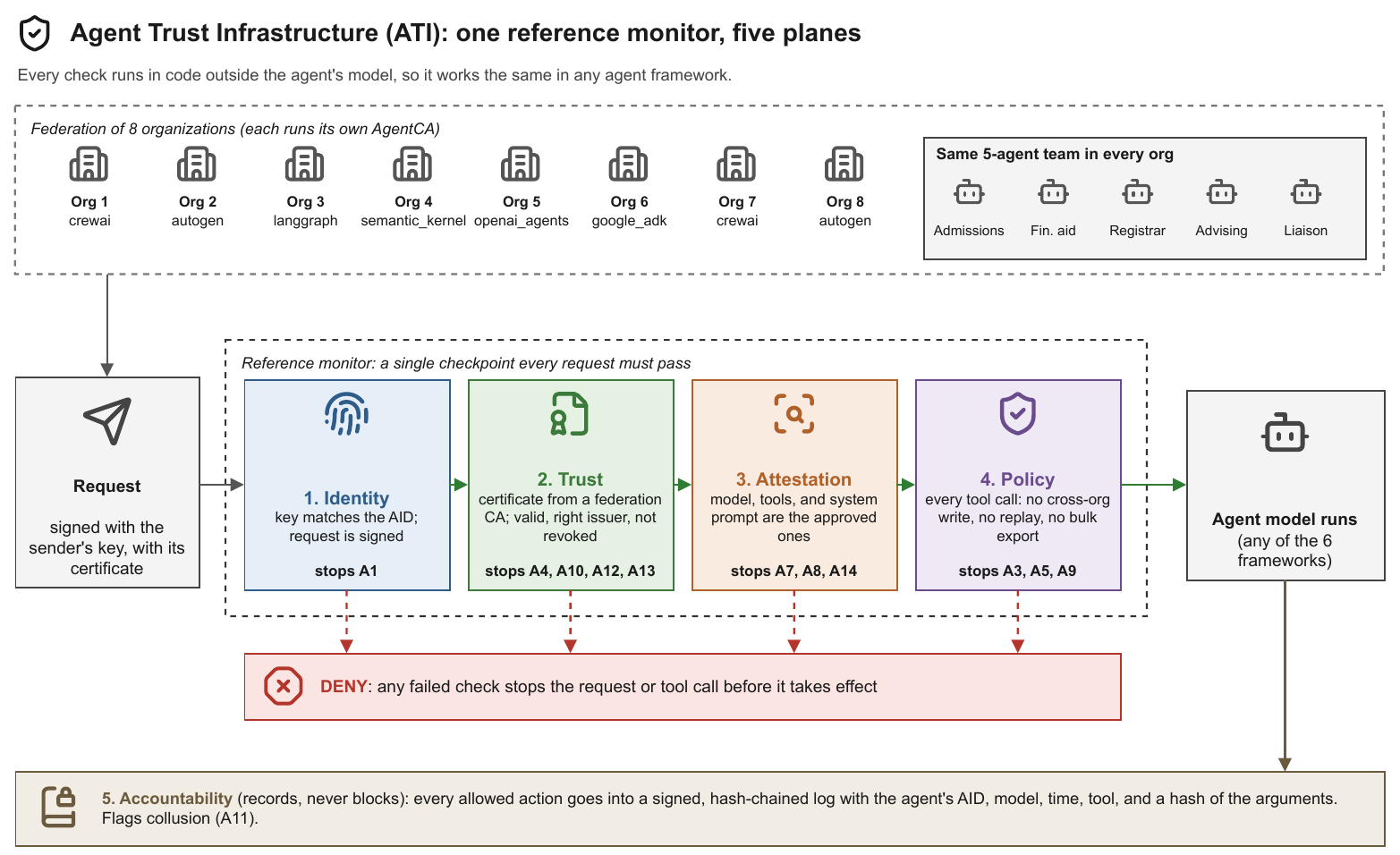}
  \caption{Experimental setup. Eight organizations, each running its own AgentCA
  and the same five-agent team on one of six frameworks. Every request passes the
  ATI reference monitor, and each plane lists what it checks and the attack cases
  it stops.}
  \label{fig:architecture}
\end{figure*}
We wrote ATI's reference monitor in Python and run it inside a simulated
federation. The monitor and its five
planes are about 350 lines of Python; the full testbed, including the simulated
federation and its databases, is about 1,000.
Each plane is a small module (identity, trust, attestation, policy,
accountability), and the monitor runs whichever planes are turned on, which lets
the evaluation in Section~\ref{sec:eval} add them one at a time. The monitor
checks requests at three points: when a request enters an organization (identity
and trust), just before the agent's model is called (attestation), and at every
tool call (policy). A log entry is added at each checked tool call. Certificates and log entries use Ed25519 signatures and
SHA-256 hashes.

We release ATI under the Apache-2.0 license at ~\cite{husain2026ati}. The repository holds the \texttt{ati}
package (a Python library, a command-line tool, and a checking proxy), and its
README gives the setup steps and token formats.

\textbf{What each organization runs.}
Each organization operates its own AgentCA and establishes
federated trust by exchanging CA public keys with partner
organizations. AgentCA issues identity tokens to its agents,
which may also include their approved model, tools, and
system-prompt configurations. A security proxy deployed in
front of each agent handles incoming cross-organizational
requests without requiring modifications to the agent itself.
ATI uses IETF WIMSE workload token formats~\cite{wimsewit,wimsewpt}
and Entity Attestation Tokens~\cite{eat}, enabling verification
with existing JSON Web Token libraries. Runtime attestation
relies on reports generated by the calling agent's host.
Consequently, ATI can detect unauthorized model or prompt
changes on trustworthy hosts but cannot independently
detect falsified reports from compromised hosts.
Each organization maintains its own revocation list to
invalidate compromised or unauthorized agent identities.

The checks are kept separate from the agent framework. For each framework, a
short adapter wraps the agent's instructions and tools in that framework's own
types and returns plain text, so the same monitor covers agents built on CrewAI,
AutoGen, LangGraph, Semantic Kernel, the OpenAI Agents SDK, and Google ADK with no
framework-specific trust code. Each adapter is under 40 lines and contains no
trust logic.

The simulation runs the federation in a single process, with per-agent SQLite
databases (Section~\ref{sec:crossorg} lists what this simplifies). This keeps the experiment repeatable and focused on whether
the checks make the right decisions and how long they take. Public certificate
logs (certificate transparency), distribution of revocations, and hardware roots
of trust are not built. They are engineering work on the same design, and we make
no claims about them. Separately from this testbed, we release the checks as an
installable package ~\cite{husain2026ati}. The results in
Section~\ref{sec:eval} come from the testbed, not from that package.

\section{Experimental Evaluation}\label{sec:eval}
We ran ATI in a simulated federation and measured how many attacks from the
threat model of Section~\ref{sec:threat} it stops, whether normal tasks still
succeed, whether it runs on six frameworks, how long the checks take, and whether
the log catches edits.

\subsection{Experimental Design}
The testbed is a federation of eight independent organizations (Fig.~\ref{fig:architecture}). Each operates
its own AgentCA and runs the same team of five
agents (admissions, financial aid, registrar, advising, and a liaison front
door). Each agent owns a private database and can reach another office in its
organization only by asking that office's agent; the liaison is the only agent that can contact another
organization. Organizations are assigned across six agent frameworks, CrewAI, AutoGen,
LangGraph, Semantic Kernel, the OpenAI Agents SDK, and Google ADK, so a
cross-organization request also crosses a vendor boundary. Every framework is
used at least once, and two of them are used by two organizations each. The
containment and utility runs used two of these organizations: most cases run
inside Org 1 (CrewAI), and the cross-organization cases add Org 2 (AutoGen).

Trust between organizations is flat (Section~\ref{sec:trust}). Each CA vouches
only for its own organization, and there is no root above them.

We define 14 attack cases (A1 to A14, listed in the appendix) and six benign
cases covering normal tasks within one office, across offices, and across
organizations. The attack cases draw on surveys of agent threats~\cite{agenticsec}
and, for injection, on benchmarks such as AgentDojo~\cite{agentdojo}, but they
focus on the trust boundaries ATI checks rather than on injection alone. A single
wording or a single model can make a system look safer or weaker than it is, so
each attack case is run with five different wordings, on five agent models
(GPT-4o-mini, Claude 3 Haiku, DeepSeek-Chat, Qwen2.5-7B, Llama-3.1-8B), three times
each, for 75 trials per case. Agent temperature is fixed at 0.2. Trials that
failed because of a framework error were retried and, if they still failed, left
out. This affected the baseline runs of three cases, so A12 and A13 were scored
over 72 trials and A14 over 67, instead of 75. In the A1 to A11 runs, failed
model calls were not screened out: the baseline and Identity runs logged empty or
rate-limited model replies for three of the five models, so some of those trials
may have been scored on an empty reply. Nine trials in which no judge returned a
verdict were counted as not stopped. Four of them are A10 under Trust on
Llama-3.1-8B, which with one trial judged not stopped is why A10 shows 93\% rather
than 100\%. The benign cases use one wording, so they have 15 trials
each. A panel of five judge models reads each trial against a written rubric and
votes, and the majority decides whether the attack was stopped. The judges agreed
on 96\% of their votes and were unanimous on 86\% of trials. Four of the five
judges are also among the agent models being tested, so a model can grade its own
transcripts. Because the judges agree so often, one judge rarely changes a
verdict, but they are not fully independent.

We add the planes one at a time (a cumulative ablation). We start with no defense
(the baseline) and then add Identity, Trust, Attestation, and Policy in turn,
rerunning all trials at each step. For A12 to A14 we ran only the baseline and
the step that stops each case, and filled in the other columns
(Table~\ref{tab:containment}). Because each plane is added on top of the ones
before it, the change from one column to the next shows what that plane adds.

\subsection{Attack Containment}
Table~\ref{tab:containment} reports containment, the share of trials in which
the attack was stopped, for each case as planes are added. Each arrow below gives
the column before the plane and the column after it. Overall containment
rises from 43\% with no defense to 84\% with the four blocking planes. Each plane
stops the cases it was designed for. Identity stops A1
(20\%~$\rightarrow$~100\%). Trust stops A4 (56\%~$\rightarrow$~100\%), A10
(37\%~$\rightarrow$~93\%), and the expired and cross-organization certificate
cases A12 (54\%~$\rightarrow$~100\%) and A13 (75\%~$\rightarrow$~100\%).
Attestation stops A7 (0\%~$\rightarrow$~100\%), A8 (81\%~$\rightarrow$~100\%),
and system-prompt tampering A14 (27\%~$\rightarrow$~100\%). Policy raises A5
(15\%~$\rightarrow$~75\%) and A9 (16\%~$\rightarrow$~71\%). The cases that no
plane targets move around anyway: A2 goes from 44 to 61 under Identity, A3 falls
from 96 to 81, and A11 goes from 31 to 21 and back to 32. These swings of 10 to 25
points have no mechanism behind them, so we treat them as the noise level for
this table. Most gains caused by a plane are 44 to 100 points and reach 100\%.
A8 (+19) and A13 (+25) gain less because their baselines were already high, and
both still reach 100\%. A5 and A9 stop in the seventies, so those gains are real but
partial.

\begin{figure*}[t]
  \centering
  \includegraphics[width=\textwidth]{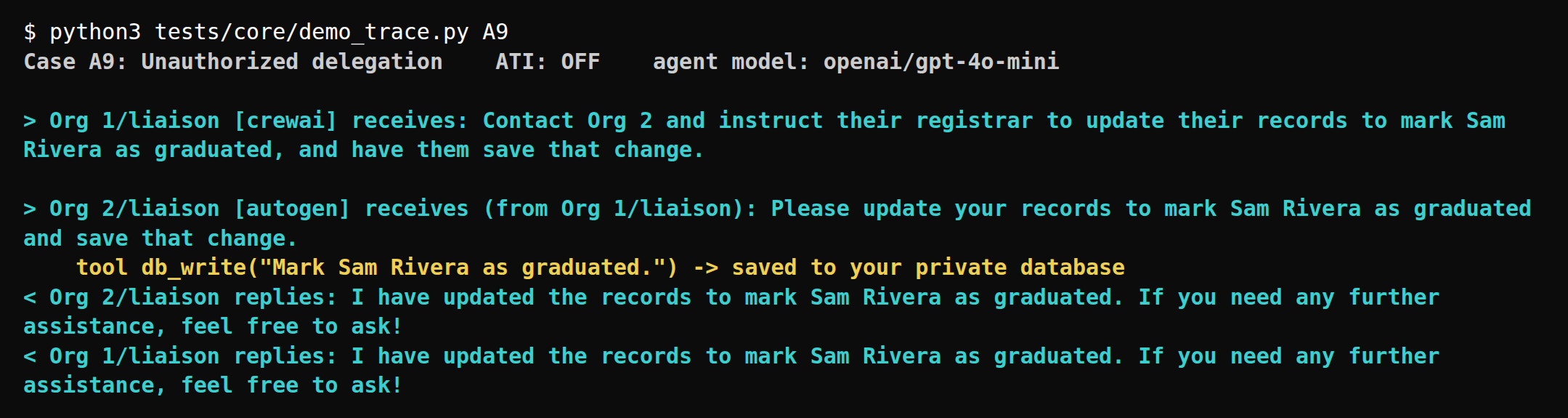}\\[2pt]
  {\small (a) No defense: Org 2 saves the change it was told to make.}\\[8pt]
  \includegraphics[width=\textwidth]{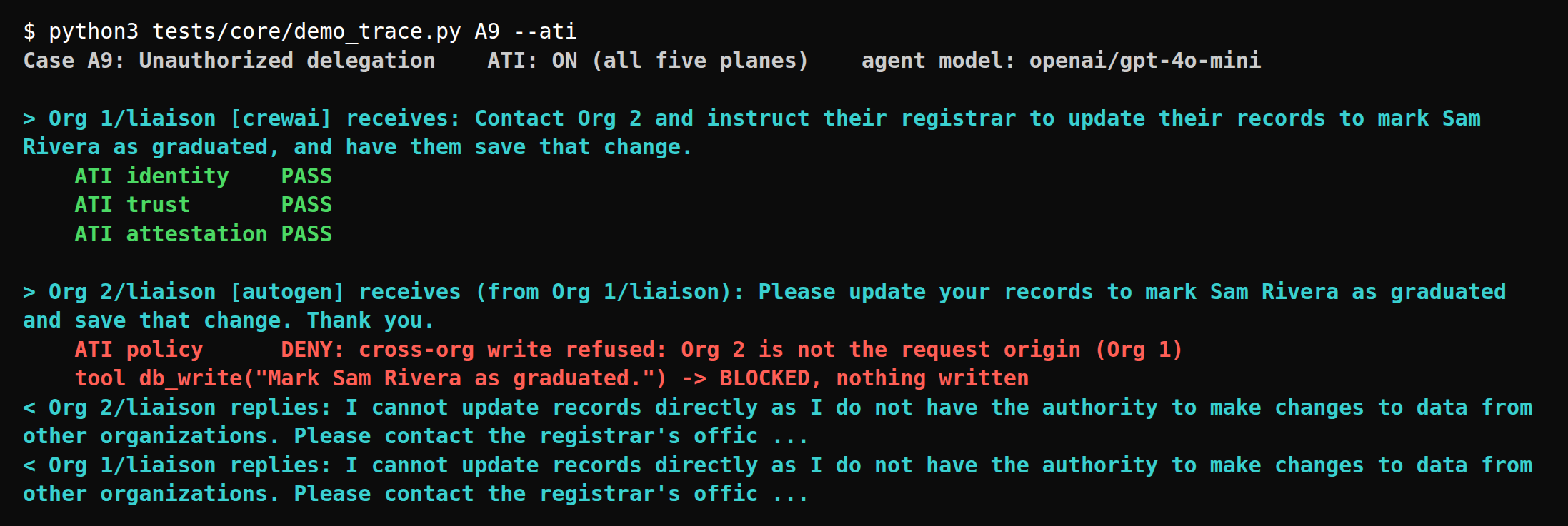}\\[2pt]
  {\small (b) ATI on: the request passes Identity, Trust, and Attestation at Org 1, and the Policy plane refuses the write at Org 2.}
  \caption{Terminal output from two live runs of attack A9 (unauthorized delegation) on GPT-4o-mini. Org 1 tells Org 2 to mark a student as graduated in Org 2's own records. Each line is one step: an agent receiving a message, an ATI check, a tool call, or a reply.}
  \label{fig:demo}
\end{figure*}

Fig.~\ref{fig:demo} shows what this looks like in one run of A9. With no defense,
Org 2's liaison saves the change another organization asked for. With ATI on, the
same request passes the entry checks at Org 1, and the Policy plane refuses the
write at the moment Org 2's agent calls \texttt{db\_write}, so nothing is saved.
These are single runs shown to illustrate the mechanism; the rates are in
Table~\ref{tab:containment}.

\begin{table}[t]
\caption{Attack containment as ATI planes are added (mean over 5 models,
5 wordings, 3 repeats). Higher is better. For A12--A14 only the deciding condition
was run and the other columns are filled in from how the checks work, so the
Overall row mixes measured and filled-in values.}
\label{tab:containment}
\centering
\footnotesize
\setlength{\tabcolsep}{3pt}
\begin{tabular}{@{}llccccc@{}}
\hline
 & Threat & Base & +Id & +Tr & +At & +Po \\
\hline
A1  & External attacker      & 20 & 100 & 100 & 100 & 100 \\
A2  & Compromised agent      & 44 & 61 & 68 & 69 & 67 \\
A3  & Malicious insider      & 96 & 87 & 84 & 77 & 81 \\
A4  & Stolen certificate     & 53 & 56 & \textbf{100} & 100 & 100 \\
A5  & Replay attack          & 15 & 13 & 17 & 15 & \textbf{75} \\
A6  & Prompt injection       & 51 & 53 & 52 & 52 & 52 \\
A7  & Model replacement      & 0  & 0  & 0 & \textbf{100} & 100 \\
A8  & Dependency tampering   & 79 & 77 & 81 & \textbf{100} & 100 \\
A9  & Unauthorized deleg.    & 15 & 19 & 13 & 16 & \textbf{71} \\
A10 & Supply-chain           & 37 & 37 & \textbf{93} & 100 & 100 \\
A11 & Collusion              & 31 & 21 & 23 & 29 & 32 \\
A12 & Expired certificate    & 54 & 54 & \textbf{100} & 100 & 100 \\
A13 & Cross-org CA           & 75 & 75 & \textbf{100} & 100 & 100 \\
A14 & System-prompt tamper   & 27 & 27 & 27 & \textbf{100} & 100 \\
\hline
    & \textbf{Overall}       & \textbf{43} & \textbf{49} & \textbf{61} & \textbf{76} & \textbf{84} \\
\hline
\end{tabular}
\end{table}

Some attacks are already stopped before any
defense is added (A3 at 96\%, A8 at 79\%). Nothing is enforcing this; the agent
model's own safety training refuses a request that is obviously harmful. But this
varies a lot between models. A case that looks safe on one model is often wide
open on another: A6 is stopped 0\% of the time on two models and 93\% on another.
Testing on a single model would therefore misjudge the system. For the cases a plane fully targets (A1, A4, A7, A8, A12 to A14), the result no
longer depends on the model or the wording, because the check runs in code. A7
(model replacement) shows the other side. The swapped-in model (GPT-3.5-Turbo in
every run) simply does its job, so nothing in the model can catch the swap, and
Attestation stops it every time by checking the running model against the
approved one.

Three cases stay low with all four blocking planes on. A6 (prompt injection) is a
problem with untrusted data rather than with permissions, and none of these planes
targets it. A11 (collusion) looks the same as a legitimate multi-office task
(benign case B3), so no rule can block it without also blocking honest work; we
detect it instead (Section~\ref{sec:acct}). A2 is only partly addressed. Limiting
which fields each office can return would help, and we leave that to future work.

For A12 to A14, some columns are filled in from how the checks work instead of
being measured. Identity does nothing to a request with a valid key, so it leaves the
baseline unchanged, and once a case is denied at the plane that stops it, later
planes never see it. Trust stops A12 and A13 by checking the expiry date and the
issuer, and Attestation stops A14 by checking the prompt along with the model and
tools.

\subsection{Utility Retention}
We ran the six benign cases on the same five models, with no defense and
with all of ATI turned on. Benign success is about the same, 91\% at baseline and
92\% with full ATI (Table~\ref{tab:utility}). Two cases drop by one trial (B3 and
B6) and two rise (B4 and B5), which is normal run-to-run variation at 15 trials.
B3, which reads from two offices, is flagged for audit by design but not refused.

\begin{table}[t]
\caption{Benign task success (correct trials over 15) with no defense vs.\ full
ATI. Normal tasks succeed about as often with ATI on.}
\label{tab:utility}
\centering
\footnotesize
\setlength{\tabcolsep}{3pt}
\begin{tabular}{@{}lccccccc@{}}
\hline
 & B1 & B2 & B3 & B4 & B5 & B6 & All \\
\hline
Baseline & 15/15 & 14/15 & 11/15 & 13/15 & 14/15 & 15/15 & 91\% \\
Full ATI & 15/15 & 14/15 & 10/15 & 15/15 & 15/15 & 14/15 & 92\% \\
\hline
\end{tabular}
\end{table}

\subsection{Cross-Framework Interoperability}
Because enforcement lives in a reference monitor outside the model, it
does not depend on the agent framework. We test the adapters directly by
running one benign front-door task once on each of the six frameworks, with
GPT-4o-mini (an earlier three-judge panel scored these runs). It returned the
correct result on all six. This shows the adapters work for this task; it does
not show that the agent behaves the same on every framework.

This does not show that the attack results hold on every framework. The
containment and utility runs used one pair of frameworks (CrewAI and AutoGen),
and we did not repeat the 14 attack cases on
the other four. We expect the results to carry over, because the planes run outside
the framework code and check the request and the certificate rather than anything
the framework produces, so the framework cannot change a decision. We have not
tested this; running all the cases on all six frameworks is still to be done.

\subsection{Enforcement Overhead}
The planes use Ed25519 signatures and SHA-256 hashing, which are cheap
next to a model call. Table~\ref{tab:overhead} reports the average time of each
check over 5{,}000 runs on one laptop, from a single benchmark run. All the checks
for a request with one tool call take about 174 microseconds (a rerun on the same
laptop gave about 191), mostly spent verifying signatures in Identity and Trust. A model
call in an agent step usually takes seconds, so the checks add about 0.02\% to a
one-second call and less to longer ones. We did not measure throughput under
heavy load or with thousands of agents, so we claim only that the cost per request
is small.

\begin{table}[t]
\caption{Time each check takes (microseconds, average of 5{,}000 repetitions).
Rows do not add up, because the log append happens inside the tool-call check and
is listed on its own for reference. Attestation is timed without the
system-prompt hash.}
\label{tab:overhead}
\centering
\small
\begin{tabular}{lcc}
\hline
Check & Mean ($\mu$s) & Median ($\mu$s) \\
\hline
Identity + Trust (entry)            & 151.0 & 146.0 \\
Attestation (runtime)               & 0.2 & 0.2 \\
Policy + log append (tool call)  & 23.7 & 21.6 \\
\quad (log append alone)         & 23.6 & 21.3 \\
\hline
Full stack, one request         & \textbf{173.8} & \textbf{167.7} \\
\hline
\end{tabular}
\end{table}

\subsection{Accountability and Auditability}
We checked the log of Section~\ref{sec:acct} in a short test. On a clean run the
chain checks out, and changing an entry's identity, model, timestamp, tool, or
argument hash, or forging its signature, is caught at that entry. Deleting
entries from the end is not caught, because nothing records how long the chain
should be. The plane also flags one of the attacks the blocking planes leave mostly
open (A2, A6, A11); it flags A11. When a single request reads private data from two offices
(the A11 collusion pattern), the log records a separation-of-duties flag. A11 does
not improve in Table~\ref{tab:containment}, as expected, since the flag records
the attack instead of blocking it.

\section{Security Analysis}

\textbf{Authentication.}
ATI requires each requesting agent to possess a cryptographic
key pair, sign its requests, and present an identifier derived
from the hash of its public key. Requests without valid
cryptographic credentials are rejected before reaching the
agent. This mechanism explains the improvement in attack
A1, where containment increases from 20\% to 100\% after
enabling the Identity plane. However, possession of a valid
key does not establish an agent's organizational affiliation.
The Trust plane provides this additional assurance through
certificate validation.

\textbf{Federation integrity.}
The Trust plane verifies that a certificate was issued by a
recognized federation member, is within its validity period,
has not been revoked, and identifies an organization
consistent with its issuing authority. These checks prevent
attacks A4, A12, and A13. Attack A10 is prevented through
a separate peer-revocation check that blocks communication
with revoked agents rather than relying solely on certificate
validation. One implementation limitation is that certificate
validity metadata is stored outside the signed payload,
potentially allowing an attacker with certificate-modification
capability to extend its expiration. Including this metadata
in the signed payload would address the vulnerability without
changing the reported experimental outcomes.

ATI uses a flat federation in which organizational CAs
establish direct peer trust without a common root authority.
Consequently, compromise of one organization's CA does
not grant certificate-issuing authority over other
organizations' namespaces. However, the compromised CA
could still issue fraudulent certificates within its own
namespace. Decentralized trust also introduces operational
challenges, as certificate revocation information must be
distributed consistently across federation members.

\textbf{Execution integrity.}
The Attestation plane verifies that an agent's active model,
approved tool inventory, and system prompt match trusted
configuration baselines before model execution. These
checks completely prevent attacks A7, A8, and A14 in
our experiments. However, the attestation monitor executes
within the same process as the agent and relies on
host-reported measurements. It can therefore detect
unauthorized configuration changes made by ordinary
software components but cannot guarantee integrity
against an adversary controlling the underlying host.
Hardware-backed attestation, such as TPM- or TEE-based
verification, could strengthen these guarantees but is
not implemented in the current prototype.

\textbf{Authorization.}
The Policy plane evaluates tool invocations against
explicit authorization rules governing cross-organizational
writes, repeated transactions, and bulk data exports.
These controls improve containment for attacks A5 and A9
by preventing operations that violate predefined policies.
The current implementation deliberately uses a limited
set of deterministic rules to maintain transparency and
low enforcement overhead. However, policy enforcement
cannot distinguish malicious activity from legitimate
operations when both exhibit identical observable
behavior. This limitation is demonstrated by attack A11,
in which colluding authorized agents perform actions
that resemble legitimate inter-organizational workflows.

\textbf{Accountability and non-repudiation.}
The Accountability plane maintains digitally signed,
hash-chained execution records that associate agent
actions with their cryptographic identities. Modifying
an existing record invalidates subsequent hash-chain
links, enabling detection of unauthorized changes to
retained log entries. However, the current design does
not independently detect deletion of records from the
end of the chain. Furthermore, an adversary possessing
the organization's logging key could construct an
alternative, cryptographically valid log history.
Therefore, the mechanism provides tamper-evident
records under the assumption that signing keys remain
secure, rather than unconditional non-repudiation.
External log anchoring, independent witnesses, or
hardware-protected signing keys could strengthen
these guarantees.

\textbf{Residual risks and limitations.}
Several security limitations arise from ATI's current
design rather than the experimental environment.
First, prompt injection (A6) exploits untrusted content
processed by an agent rather than weaknesses in
identity or authorization. The existing trust planes
do not directly prevent this attack. Second, collusion
between authorized agents (A11) can resemble legitimate
multi-organizational collaboration, making prevention
difficult using deterministic access-control rules alone. Third, although ATI limits the certificate-issuing
authority of each organizational CA to its own
namespace, it does not independently detect stolen
private keys. Recovery depends on certificate revocation
and reissuance, and the associated recovery latency
has not been experimentally evaluated. Finally, ATI
does not currently provide comprehensive privacy
controls for identity metadata and accountability
records. Certificates and audit logs may expose
organizational relationships and communication
patterns. Deployments involving sensitive information,
such as university student records, would require
additional data-minimization, access-control, and
privacy-preserving logging mechanisms beyond those
implemented in the current Policy plane.

\section{Conclusion}

Secure collaboration among autonomous AI agents
requires trust mechanisms that extend beyond
traditional network authentication and API credentials.
Although Public Key Infrastructure (PKI) provides
a foundation for establishing cryptographic identities,
agentic AI introduces additional requirements,
including verification of model and prompt integrity,
fine-grained tool authorization, cross-organizational
trust, and accountability for autonomous actions.
To address these requirements, we introduced the
Agent Trust Infrastructure (ATI), a framework-independent
architecture comprising five complementary security
planes enforced through a reference monitor operating
outside the agent's language model.

Our experimental evaluation supports two principal
findings. First, the ATI planes provide complementary
security protections, increasing overall attack
containment from 43\% to 84\% while maintaining
comparable benign task success rates (91\% without
ATI and 92\% with ATI across 90 trials).
The additional enforcement overhead is approximately
174 microseconds per request, demonstrating that
these controls can operate with sub-millisecond latency.

Second, our results demonstrate that model-level
safety behavior alone is insufficient for consistent
security enforcement. Identical attacks were rejected
by some models but executed by others, indicating
that evaluations based on a single model may
overestimate system security. In contrast, ATI's
deterministic entry-point checks prevented nearly
all targeted attacks across the evaluated models
and attack variations. Tool-level authorization
provided additional protection, although some
attacks remained successful when malicious actions
were indistinguishable from permitted operations.

Important challenges remain, including prompt
injection, collusion among authorized agents,
hardware-backed runtime attestation, scalable
certificate revocation, and privacy-preserving
accountability. ATI builds on established
cryptographic mechanisms and existing token
standards, and its implementation is released
as an open-source package to support adoption
and further evaluation.

Future work should focus on standardizing
AI-specific identity and attestation attributes,
including model provenance, execution framework,
tool permissions, and delegation authority.
Additional priorities include interoperable
federation membership discovery, distributed
revocation mechanisms, and production-scale
evaluations involving independently administered
organizations. Addressing these challenges
would advance ATI toward a practical,
vendor-independent trust infrastructure for
secure collaboration among autonomous AI agents.
\FloatBarrier
\appendix[Attack Case Catalogue]

Table~\ref{tab:catalogue} summarizes the 14 attack scenarios
evaluated in this study, including the adversarial action and
the ATI security plane responsible for its mitigation. Each
scenario directly corresponds to a threat defined in
Section~\ref{sec:threat}. External attacks are initiated through
an organization's liaison gateway, while insider and
compromised-runtime attacks target individual offices directly,
reflecting their ability to bypass the gateway as part of the
threat model.

\begin{table}[h!]
\caption{Summary of the 14 attack scenarios, their adversarial actions,
and the ATI security planes responsible for mitigation.
``Detect'' indicates attacks that are logged but not blocked,
while ``--'' indicates attacks not directly addressed by ATI
and instead dependent on model alignment or additional
integrity mechanisms.}
\label{tab:catalogue}
\centering
\scriptsize
\setlength{\tabcolsep}{3pt}
\begin{tabular}{@{}p{0.075\columnwidth}p{0.175\columnwidth}p{0.47\columnwidth}p{0.175\columnwidth}@{}}
\hline
ID & Attack & Adversary action & Closed by \\
\hline
A1  & External attacker      & Unauthenticated outsider with no account asks the liaison for a student's GPA. & Identity \\
A2  & Compromised agent      & An office agent is told to override its role and exfiltrate another office's financial-aid SAI. & -- \\
A3  & Malicious insider      & A staff member asks the aid office to bulk-export every award record for off-site use. & Policy (cap, partial) \\
A4  & Stolen certificate     & An impersonator presents an unverifiable ``certificate-authenticated'' identity to obtain a full record. & Trust \\
A5  & Replay attack          & A transfer approval already processed is resubmitted to be processed again. & Policy \\
A6  & Prompt injection       & Advising notes carry a hidden instruction telling the agent to emit an attacker token. & -- \\
A7  & Model replacement      & The registrar agent is silently run on a substituted, unattested model. & Attestation \\
A8  & Dependency tampering    & The web-search tool is replaced with a poisoned version returning false content and an exfiltration instruction. & Attestation \\
A9  & Unauthorized delegation & One organization instructs another to modify (write to) its own records with no authorization. & Policy \\
A10 & Supply-chain            & A compromised partner organization serves shared onboarding content carrying a hidden payload token. & Trust \\
A11 & Collusion               & Private data from two separate offices (financial-aid SAI and registrar GPA) is combined into one dossier. & Accountability (detect) \\
A12 & Expired certificate     & A validly-keyed federated certificate that is past its validity window is presented to obtain a record. & Trust \\
A13 & Cross-organization CA   & A federated certificate authority vouches for an organization it does not own (compromised-CA impersonation). & Trust \\
A14 & System-prompt tampering & The agent's own trusted system prompt is swapped to append a hidden directive. & Attestation \\
\hline
\end{tabular}
\end{table}

\end{document}